\documentclass{article}
\usepackage{times}
\usepackage{ijcai16}
\usepackage{mathrsfs}
\usepackage{amsmath}
\usepackage{bm}%在数学公式下加黑字体
\usepackage{clrscode}
\usepackage{graphicx}
\usepackage{amsfonts, amssymb}
\usepackage{url}
\usepackage{algorithm}
\usepackage{algorithmic}
\usepackage{multirow}
\usepackage{balance}
\usepackage{fancybox, xcolor}

\usepackage{pifont}

\newtheorem{lem}{Lemma}
\newtheorem{ex}{Example}

\usepackage{tikz}
\usetikzlibrary{shapes, arrows, positioning, decorations.markings}

\definecolor {processblue}{cmyk}{0.96,0,0,0}
\tikzstyle{int}=[draw, fill=blue!20, minimum size=2em]
\tikzstyle{init} = [pin edge={to-,thin,black}]

\usetikzlibrary{shapes}
\usetikzlibrary{fit}
\usetikzlibrary{chains}
\usetikzlibrary{arrows}

\tikzstyle{plate} = [draw, rectangle, rounded corners, fit=#1]
\tikzstyle{wrap} = [inner sep=0pt, fit=#1]

\tikzstyle{caption} = [node distance=0] %
\tikzstyle{bottom plate caption} = [caption, node distance=0, inner sep=0pt,
below left=-5pt and 0pt of #1.south east] %

\tikzstyle{top plate caption} = [caption, node distance=0, inner sep=0pt,
below left=0pt and 0pt of #1.north east] %

\title{On the Representation and Embedding of Knowledge Bases\\
 Beyond Binary Relations}
\author{Jianfeng Wen$^{1,2}$, Jianxin Li$^{1,2}$, Yongyi Mao$^3$, Shini Chen$^1$, \and Richong Zhang\thanks{Corresponding author: zhangrc@act.buaa.edu.cn}$^{1,2}$\\
$^1$ State Key Laboratory of Software Development Environment, Beihang University\\
$^2$ School of Computer Science and Engineering, Beihang University\\
$^3$ School of Electrical Engineering and Computer Science, University of Ottawa}

\begin{document}

\maketitle

\begin{abstract}

The models developed to date for knowledge base embedding are all based on the assumption that the relations contained in knowledge bases are binary.
%, namely, that each relational instance involves only two entities.
For the training and testing of these embedding models, multi-fold (or n-ary) relational data are  converted to triples (e.g., in FB15K dataset) and interpreted as instances of binary relations.
%
%
%Under such an assumption, the data in a knowledge base are first converted to entity-predicate-entity triples,  and all existing embedding models are developed for data in such a triple-based representation.  As knowledge bases usually contain a significant amount non-binary relational data, we take a fundamental look at the representation and embedding of
This paper presents a canonical representation of knowledge bases containing multi-fold relations. We show that the existing embedding models on the popular FB15K datasets correspond to a sub-optimal modelling framework, resulting in a loss of structural information. We advocate a novel modelling framework, which models multi-fold relations directly using this canonical representation. Using this framework, the existing TransH model is generalized to a new model,  m-TransH. We demonstrate experimentally that  m-TransH outperforms TransH by a large margin, thereby establishing a new state of the art.

\end{abstract}

\section{Introduction}
The emerging of knowledge bases such as YAGO\cite{suchanek2007yago}, DBpedia\cite{auer2007dbpedia} and Freebase\cite{bollacker2008freebase} has inspired intense research interest in this area, from completing and improving knowledge bases (e.g., \cite{kgCompleting2007,kgCompleting2013}) to developing applications that retrieve information from the knowledge data (e.g., \cite{KG-APP1,KG-APP2,KG-APP3}). Recently knowledge base embedding\cite{transE,transH,transR,UE_SME,SE,SLM,PtransE} has stood out as an appealing and generic methodology to access various research problems in this area.  Briefly, this methodology sets out to represent entities in a knowledge base as points in some Euclidean space while preserving the structures of the relational data. This approach turns the discrete topology of the relations into a continuous one, enabling the design of efficient algorithms and potentially benefitting many applications. For example, embedding can be applied to link prediction \cite{transE} or question answering \cite{qasubg2014} in knowledge bases, in which the problems are usually of a combinatorial nature in their original discrete settings.

Despite their promising successes, existing embedding techniques are all developed based on the assumption that knowledge data are instances of binary relations, namely instances each involving two entities (such as ``Beijing is the capital of China"). In reality, however, a large portion of the knowledge data are from non-binary relations (such as ``Benedict Cumberbatch played Alan Turing in the movie The Imitation Game'').  For example, we observe that in Freebase\cite{bollacker2008freebase}, more than 1/3 of the entities participate in non-binary relations.  This calls for a careful investigation of embedding techniques for knowledge bases containing non-binary relations.

In this paper, we first present a clean mathematical definition of {\em multi-fold relations}, also known as n-ary relation in the literature \cite{nAryDB}. Using this notion, we propose a canonical representation for multi-fold
(binary or non-binary) relational data, which we call {\em instance representation}. Existing knowledge bases usually organize their data using the W3C Resource Description Framework (RDF)~\cite{Google2015Review}, in which relational data are represented as a collection of (subject, predicate, object) triples. Although such a {triple representation}, like the instance representation,   is capable of capturing the structures of multi-fold relations~\cite{nAryDB,KGNARY-1,NARY-IMP1,NARY-IMP2,NARY-IMP3,W3C-nary}, we show that manipulating multi-fold relational data into triples (as in Freebase) results in an heterogeneity of the predicates, unfavourable for embedding. As such, we advocate that the starting point of embedding multi-fold relations should be recovering the relational data in its instance representation.

We then formulate the embedding problem on the instance representation  and suggest that the heart of the problem is modelling each cost function that defines a constraint in the embedding space.  We examine the popular  FB15K\cite{transE} datasets, used in all existing embedding models, and point out that the triple-based data format of FB15K results from applying a particular ``star-to-clique" (S2C) conversion procedure to the filtered Freebase data. This procedure can be verified to be irreversible, which causes a loss of structural information in the multi-fold relations.

Interestingly, we discover that all existing embedding models on such S2C-converted datasets can be unified under a ``decomposition" modelling framework. In this framework, the cost function associated with a $J$-fold relation is modelled as the sum of ${J \choose 2}$ bi-variate functions. Suggesting that the decomposition framework is fundamentally limited, we propose a ``direct modelling" framework for embedding multi-fold relations. As an example in this framework, we generalize TransH\cite{transH} to a new model for multi-fold relations. Although TransH is known to perform
comparably to the best performing models
but with lower time complexity,  our experiments demonstrate that this new model, designated {\em m-TransH}, has even lower complexity, and outperforms TransH by an astonishing margin.

In summary, this paper takes a fundamental look at the knowledge base embedding problem when non-binary relations exist.  We  advocate the instance representation and  the direct modelling framework on such representation.  Constrained by the length requirement, we are unable to elaborate at places and certain details are omitted.

%\vspace{-0.3cm}
\section{Multi-Fold Relations and Knowledge Base Representations}
%\vspace{-0.5ex}

\subsection{Multi-Fold Relations}

%\vspace{-0.2ex}

A well-known algebraic concept,  a binary relation \cite{hungerford} on a set ${\cal N}$ is defined as a subset of the cartesian product ${\cal N}\times {\cal N}$, or ${\cal N}^2$.  This understanding allows an immediate generalization of binary relation to multi-fold relation (also known as n-ary relation, see, e.g. \cite{nAryDB})  where ${\cal N}^2$ is replaced with the $J$-fold cartesian product ${\cal N}^J$ for an arbitrary integer $J\ge 2$. From a knowledge base (KB) point of view, we argue however that such algebraic definitions are incomplete, in the sense that the {\em role} of each coordinate in the cartesian product is not specified.  For example, let $R$ be the binary relation relating a country with its capital city. Then  ambiguity exists in whether the instance ``Paris is the capital of France'' should be written as $({\rm Paris}, {\rm France}) \in R$ or as $({\rm France}, {\rm Paris}) \in R$.  Although this issue is usually resolved by suitable data structures, we now formulate a clean mathematical notion of multi-fold relation which also specifies the roles of involved entities.

Throughout the paper, the following notations will be used.  For any set ${\cal A}$ and ${\cal B}$, we denote by ${\cal B}^{\cal A}$ the set of all {\em functions} mapping ${\cal A}$ to ${\cal B}$, as is standard in mathematics  \cite{hungerford}. For any function $g\in {\cal B}^{\cal A}$ and any subset ${\cal S}\subseteq {\cal A}$, we use $g_{:{\cal S}}$ to denote the restriction of function $g$ on ${\cal S}$. We will use ${\cal N}$ to denote the set of all entities in a KB.

A multi-fold relation is then defined as follows. Let ${\cal M}$ be a set of {\em roles} in the KB, and a {\em multi-fold relation}, or simply, {\em relation},  $R$ on ${\cal N}$ with roles ${\cal M}$ is a subset of ${\cal N}^{\cal M}$. Given $R$, we also write the set ${\cal M}$ as ${\cal M}(R)$ and call $R$ a $J$-fold (or $J$-ary) relation if $|{\cal M}(R)|=J$.  The value $J$ is also referred to as the ``fold'' or ``arity'' of $R$. An element $t\in R$ is called an {\em instance} of the relation $R$.

\begin{ex} Let $R$ be a $3$-fold relation about ``which actor played which character in which movie''. Then
$
{\cal M} (R): = \{
{\rm ACTOR}, {\rm CHARACTER}, {\rm MOVIE}
\}.
$
The function $t: {\cal M}(R)\rightarrow {\cal N}$ given below is then the instance of $R$ stating ``Benedict Cumberbatch played Alan Turing in the movie The Imitation Game":
\begin{equation*}
\label{eq:instEx}
\begin{array}{lcl}
t({\rm ACTOR}) & = & {\rm BenedictCumberbatch},\\
t({\rm CHARACTER }) & = &  {\rm AlanTuring},\\
t({\rm MOVIE }) & = &  {\rm TheImitationGame}.
\end{array}
\end{equation*}
\end{ex}

%\vspace{-0.2ex}

\subsection{Instance Representation}
%\vspace{-0.2ex}
Let ${\cal R}$ index a set of distinct multi-fold relations on ${\cal N}$. More precisely put, for each $r\in {\cal R}$, there is a relation $R_r$ on ${\cal N}$ with roles ${\cal M}(R_r)$.  One may identify the index $r$ of $R_r$ with  the {\em type}  of the relation $R_r$, and in this view, ${\cal R}$ is a collection of {\em relation types}. Let ${\cal T}_r$ be the set of instances of relation ${R}_r$ that are included in the KB, then the KB can be specified as $({\cal N}, {\cal R}, \{{\cal T}_r, r\in {\cal R}\})$.  We call such specification an {\em instance representation}. Since a KB is usually incomplete,  each set ${\cal T}_r$ is expected to be strictly contained in $R_r$. As a consequence, relation $R_r$ is in fact unknown, and all information about $R_r$ is  revealed via the set  ${\cal T}_r$,  sampled from $R_r$. Clearly, the instance representation contains all information in the KB pertaining to the structures of the relations and how they interact.  Therefore, instance representations are legitimately canonical, at least from the embedding perspective, where only such information matters.
%\vspace{-0.2ex}

\subsection{Fact Representation}
\label{subsec:metaRelation}

%\vspace{-0.2ex}

For various implementation considerations, practical KBs such as Freebase \cite{FreebaseWeb} organize relational data in a different format, the core of which is a notion related to but different from the multi-fold relation we define. We call this notion {\em meta-relation} and define it next.

Given a set ${\cal N}$ of entities and a set ${\cal M}$ of roles,  a {\em (multi-fold) meta-relation} $Q$ on ${\cal N}$ with roles ${\cal M}$ is a subset of $\left(2^{\cal N}\right)^{\cal M}$, where $2^{\cal N}$ is the power set of ${\cal N}$. That is, each element in the meta-relation $Q$, which will be called a {\em fact} of $Q$,  is a function mapping ${\cal M}$ to $2^{\cal N}$.  Similar to relations, we call $Q$ a $J$-fold meta-relation if $|{\cal M}|=J$ and often write ${\cal M}(Q)$ in place of ${\cal M}$.

\begin{ex}
\label{ex:meta-relation}
Let $Q$ be a $3$-fold meta-relation about ``who played what instruments in the recording of what music piece".
The roles ${\cal M}(Q)$ consists of
 $\rm{RECORDING}$ (i.e., the music piece), $\rm{INSTRUMENT\!\!\!-\!\!\!ROLE}$
 (i.e., the music instrument), and $\rm{CONTRIBUTOR}$ (i.e., the person).

Let $u_1$ and $u_2$ be two facts of $Q$, which respectively
state ``Will McGregor played bass and bass guitar in Precious Things" and ``Michael Harrison played violin in Pretty Good Year".   As functions in $(2^{\cal N})^{{\cal M}(Q)}$, $u_1$ and $u_2$ are given in Table \ref{tab:fact2inst}.

\end{ex}
For any given meta-relation $Q$, the number of entities involved in a fact $u\in Q$
may vary with $u$ in general.   Recall that in a $J$-fold relation,  each instance involves exactly $J$  entities.   For later reference (in Section \ref{subsec:freebase}), a meta-relation $Q$ is said to be {\em degenerate} if
every $u\in Q$ is such that $u(\rho)$ is a singleton set for every $\rho \in {\cal M}(Q)$.

Parallel to instance representations, one can similarly define the {\em fact representation} $({\cal N}, {\cal Q}, \{{\cal U}_q:q\in {\cal Q}\})$ for a KB: ${\cal Q}$ indexes a set $\{{Q}_{q}: q\in {\cal Q}\}$ of distinct meta-relations where each $Q_q$ is a meta-relation on ${\cal N}$ with roles ${\cal M}(Q_q)$;  and each ${\cal U}_q$  is a set of facts of $Q_q$.

%\vspace{-0.2ex}

\subsection{Graphical Representations}
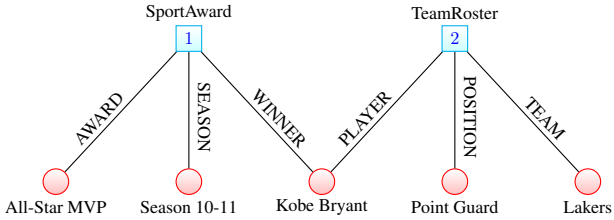
\begin{figure}[ht!]
\begin{center}
\scalebox{0.68}{
\begin {tikzpicture}[-latex ,auto ,node distance =2.8 cm and 2.6cm ,on grid ,
semithick ,
state/.style ={ circle ,top color =white , bottom color = processblue!20 ,
draw, processblue , text=blue , minimum width =0.5cm},
statebox/.style ={rectangle ,top color =white , bottom color = processblue!20 ,
draw, processblue , text=blue , minimum width =0.5 cm , minimum height =0.5 cm}]

\node[statebox] (inst_1){$1$};
\node [] (SportAwardText) [above =0.5cm of inst_1]{SportAward};

\node[state, red, bottom color=pink] (season)[below=of inst_1]  {};
\node [] (SeasonText) [below =0.5cm of season]{Season 10-11};

\path (inst_1) edge [-] node [midway, rotate=-90]  (anchor_1){}  (season);
\node [right=0.3cm of anchor_1, fill=white, rotate=-90] {SEASON};

\node[state, red, bottom color=pink] (MVP)[left=of season]  {};
\node [] (MVPText) [below =0.5cm of MVP]{All-Star MVP};

\path (inst_1) edge [-] node [midway, rotate=-90]  (anchor_2){}  (MVP);
\node [left=0.3cm of anchor_2, rotate=47] {AWARD};

\node[state, red, bottom color=pink] (Kobe)[right=of season]  {};
\node [] (KobeText) [below =0.5cm of Kobe]{Kobe Bryant};

\path (inst_1) edge [-] node [midway, rotate=-90]  (anchor_3){}  (Kobe);
\node [right=0.3cm of anchor_3, rotate=-47] {WINNER};

\node[state, red, bottom color=pink] (PointGuard)[right=of Kobe]  {};
\node [] (PointGuardText) [below =0.5cm of PointGuard]{Point Guard};

\node[statebox] (inst_2)[above = of PointGuard]{$2$};
\node [] (TeamRosterText) [above =0.5cm of inst_2]{TeamRoster};

\path (inst_2) edge [-] node [midway, rotate=-90]  (anchor_4){}  (Kobe);
\node [left=0.3cm of anchor_4, rotate=47] {PLAYER};

\path (inst_2) edge [-] node [midway, rotate=-90]  (anchor_5){}  (PointGuard);
\node [right=0.3cm of anchor_5, fill=white, rotate=-90] {POSITION};

\node[state, red, bottom color=pink] (Lakers)[right=of PointGuard]  {};
\node [] (LakersText) [below =0.5cm of Lakers]{Lakers};

\path (inst_2) edge [-] node [midway, rotate=-90]  (anchor_6){}  (Lakers);
\node [right=0.3cm of anchor_6, rotate=-47] {TEAM};

\end{tikzpicture}
}
\end{center}
\caption{The bipartite graph for a toy instance representation containing two instances. Instance 1 is  from the SportAward relation (or having relation type SportAward), stating ``Kobe Bryant is the All-Star MVP for the season 2010-2011". Instance 2 is from the TeamRoster relation, stating ``Kobe Bryant is a Point Guard in Los Angeles Lakers''.}
\label{fig:instKobe}
\end{figure}

\begin{figure}[ht!]
\begin{center}
\scalebox{0.68}{
\begin {tikzpicture}[-latex ,auto ,node distance =2.8 cm and 2.6cm ,on grid ,
semithick ,
state/.style ={ circle ,top color =white , bottom color = processblue!20 ,
draw, processblue , text=blue , minimum width =0.5cm},
statebox/.style ={rectangle ,top color =white , bottom color = processblue!20 ,
draw, processblue , text=blue , minimum width =0.5 cm , minimum height =0.5 cm}]

\node[statebox] (inst_1){$1$};
\node [] (SportAwardText) [above =0.5cm of inst_1]{PeopleMarriage};

\node[state, red, bottom color=pink] (season)[below=of inst_1]  {};
\node [] (SeasonText) [below =0.5cm of season]{Venessa Bryant};

\path (inst_1) edge [-] node [midway, rotate=-90]  (anchor_1){}  (season);
\node [right=0.3cm of anchor_1, fill=white, rotate=-90] {SPOUSE};

\node[state, red, bottom color=pink] (MVP)[left=of season]  {};
\node [] (MVPText) [below =0.5cm of MVP]{Dana Point};

\path (inst_1) edge [-] node [midway, rotate=-90]  (anchor_2){}  (MVP);
\node [left=0.3cm of anchor_2, rotate=47] {LOCATION};

\node[state, red, bottom color=pink] (Kobe)[right=of season]  {};
\node [] (KobeText) [below =0.5cm of Kobe]{Kobe Bryant};

\path (inst_1) edge [-] node [midway, rotate=-90]  (anchor_3){}  (Kobe);
\node [right=0.3cm of anchor_3, rotate=-47] {SPOUSE};

\node[] (PointGuard)[right=of Kobe]  {};
%\node [] (PointGuardText) [below =0.5cm of PointGuard]{Point Guard};

\node[statebox] (inst_2)[above = of PointGuard]{$2$};
\node [] (TeamRosterText) [above =0.5cm of inst_2]{PlaceOfBirth};

\path (inst_2) edge [-] node [midway, rotate=-90]  (anchor_4){}  (Kobe);
\node [left=0.3cm of anchor_4, rotate=47] {PERSON};

%\path (inst_2) edge [-] node [midway, rotate=-90]  (anchor_5){}  (PointGuard);
%\node [right=0.3cm of anchor_5, fill=white, rotate=-90] {POSITION};

\node[state, red, bottom color=pink] (Lakers)[right=of PointGuard]  {};
\node [] (LakersText) [below =0.5cm of Lakers]{Philadelphia};

\path (inst_2) edge [-] node [midway, rotate=-90]  (anchor_6){}  (Lakers);
\node [right=0.3cm of anchor_6, rotate=-47] {PLACE};

\end{tikzpicture}
}
\end{center}
\caption{The bipartite graph for a toy fact representation containing two facts. Fact 1 is  from the PeopleMarriage meta-relation (noting that two entities have the same role SPOUSE), stating ``Kobe Bryant and Venessa Bryant became married
in Dana Point''.
Fact 2 is from the PlaceOfBirth meta-relation, stating ``Kobe Bryant was born in Philadelphia'' (noting that such a fact is also an instance).}
\label{fig:factKobe}
\end{figure}
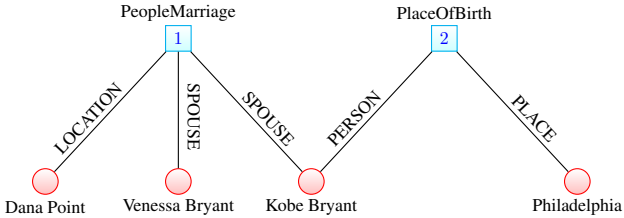

%\vspace{-0.2ex}
Both instance representations and fact representations can be associated with an edge-labelled bi-partite graph. For an instance (resp. fact) representation, its associated graph contains two sets of vertices, representing entities and instances (resp. facts) respectively; if an entity is involved in an instance (resp. fact), the corresponding entity vertex is connected to corresponding instance (resp. fact) vertex by an edge, and the edge label is the role of the entity in the instance (resp. fact). The reader is referred to Figures \ref{fig:instKobe} and \ref{fig:factKobe} for examples of such graphical notations. As such, we may sometimes use a graph-theoretic language when speaking of instance or fact representations.

%\vspace{-0.2ex}

\subsection{Converting Facts to Instances}

%\vspace{-0.2ex}

We now show that a fact representation ${\cal F}:=({\cal N}, {\cal Q}, \{{\cal U}_q: q\in {\cal Q}\})$ can be converted to an instance representation.

Let ${\cal N'}: = {\cal N}\cup \left(
\bigcup_{q\in {\cal Q}}{\cal U}_q
\right)$  and let ${\cal M'}(Q): = {\cal M}(Q) \cup \{{\rm FACT\!\!\!-\!\!ID}\}$ for any arbitrary meta-relation $Q$ on ${\cal N}$.  That is, the ID (or name) of each fact in ${\cal F}$ is regarded as an ``entity'' and an additional ``role'' ${\rm FACT\!\!\!-\!\!ID}$ is generated. For any meta-relation $Q$ on ${\cal N}$ and a fact $u\in Q$,  let $T_{\rm id}(u)$  be the set of all  distinct functions in
${\cal N'}^{{\cal M}'(Q)}$ in which each function $t$ maps a role $\rho\in {\cal M}(Q)$ to an entity in $u(\rho)$ and maps ${\rm FACT\!\!\!-\!\!ID}$ to the ID of the fact $u$. It is easy to see that
$T_{\rm id}(u)$ contains precisely $\left|\prod_{\rho\in {\cal M}(Q)} u(\rho)\right|$ functions.
%For the fact $u_1$ in Example \ref{ex:meta-relation}, $T_{\rm id}(u_1)$ consists of functions $t'_1$ and $t'_2$ in Table \ref{tab:fact2inst}.
For any collection ${\cal U}$ of facts, let $T_{\rm id}({\cal U}): =\bigcup_{u\in {\cal U}} T_{\rm id}(u)$. Then for any meta-relation $Q$, $T_{\rm id}(Q)$ is a subset of ${{\cal N}'}^{{\cal M}'(Q)}$, namely a relation on ${\cal N}'$ with roles ${\cal M}'(Q)$.

Denote $T_{\rm id}({\cal F}): = ({\cal N}', {\cal Q}, \{T_{\rm id}({\cal U}_q): q\in {\cal Q}\})$. Clearly, $T_{\rm id}({\cal F})$ is the instance representation of a KB  ``augmented''  from ${\cal F}$, where fact ID's in ${\cal F}$ are added to the entities, and ${\rm FACT\!\!\!-\!\!ID}$ is taken as an additional role.

%\vspace{-0.1cm}

\begin{lem}
\label{lem:tid}
${\cal F}$ can be recovered from $T_{\rm id}({\cal F})$.
\end{lem}
If one only wishes for an instance representation of ${\cal F}$ without demanding the recoverability in Lemma \ref{lem:tid}, a simpler conversion can be defined: for a fact $u\in Q$, $T(u): = \{t_{:{\cal M}(Q)}: t\in T_{\rm id}(u)\}$. That is, $T(u)$ is similar to $T_{\rm id}$ but having fact ID dropped. Similarly
for any collection ${\cal U}$ of facts, let $T({\cal U}): =\bigcup_{u\in {\cal U}} T(u)$.  Then for any meta-relation $Q$, $T(Q)$ is a relation on ${\cal N}$ with roles ${\cal M}(Q)$. Denote $T({\cal F}): = ({\cal N}, {\cal Q}, \{T({\cal U}_q): q\in {\cal Q}\})$. Then $T({\cal F})$ is an instance representation for  ${\cal F}$, but ${\cal F}$ can not be recovered from $T({\cal F})$ in general.

\begin{figure}[ht!]
\begin{center}
\begin{tabular}{c@{\hspace{0.2cm}}c}

%\begin{figure}[ht!]
%\begin{center}
\scalebox{0.72}{
\begin {tikzpicture}[-latex ,auto ,node distance =3.4 cm and 1.6cm ,on grid ,
semithick ,
state/.style ={ circle ,top color =white , bottom color = processblue!20 ,
draw, processblue , text=blue , minimum width =0.5cm},
statebox/.style ={rectangle ,top color =white , bottom color = processblue!20 ,
draw, processblue , text=blue , minimum width =0.5 cm , minimum height =0.5 cm}]

\node[state, black, bottom color=gray] (u1){$u_1$};

\node[state, red, bottom color=pink] (bassGuitar)[left= of u1]{};
\node[] (bassGuitarText) [below= 0.5cm of bassGuitar]{Bass Guitar};

\node[state, red, bottom color=pink] (bass)[right= of u1]{};
\node[] (bassText) [below= 0.5cm of bass]{Bass};

\node[statebox] (t1)[above=of bassGuitar]{$t'_1$};
\node[statebox] (t2)[above=of bass]{$t'_2$};

\node[state, red, bottom color=pink] (will)[above= of t1]{};
\node[] (willText) [above= 0.5cm of will]{Will McGregor};

\node[state, red, bottom color=pink] (precious)[above= of t2]{};
\node[] (preciousText) [above= 0.5cm of precious]{Precious Things};

\path (will) edge [-] node [midway]  (anchor_1){}  (t1);
\node [left =0.35cm of anchor_1, rotate=90] {CONTRIB};

\path (precious) edge [-] node [midway]  (anchor_2){}  (t2);
\node [left =0.35cm of anchor_2, rotate=90] {RECORD};

\path (t1) edge [-] node [midway]  (anchor_3){}  (bassGuitar);
\node [left =0.35cm of anchor_3, rotate=90] {INSTR-ROLE};

\path (t2) edge [-] node [midway]  (anchor_4){}  (bass);
\node [left =0.35cm of anchor_4, rotate=90] {INSTR-ROLE};

\path (t1) edge [-] node [midway]  (anchor_5){}  (u1);
\node [right =0.1cm of anchor_5, rotate=-65] {FACT-ID};

\path (t2) edge [-] node [midway]  (anchor_6){}  (u1);
\node [above left=0.3cm and 0.3cm of anchor_6, rotate=65] {FACT-ID};

\path (will) edge [-] node [midway]  (anchor_7){}  (t2);
\node [above right =0.7cm and -0.6cm of anchor_7, rotate=-47] {CONTRIB};

\path (precious) edge [-] node [midway]  (anchor_8){}  (t1);
\node [below right = 0.65cm and -0.55cm of anchor_8, rotate=47] {RECORD};

\end{tikzpicture}
}
\hspace{-.51cm}
 &
%\begin{figure}[ht!]
%\begin{center}
\scalebox{0.72}{
\begin {tikzpicture}[-latex ,auto ,node distance =3.4 cm and 1.6cm ,on grid ,
semithick ,
state/.style ={ circle ,top color =white , bottom color = processblue!20 ,
draw, processblue , text=blue , minimum width =0.5cm},
statebox/.style ={rectangle ,top color =white , bottom color = processblue!20 ,
draw, processblue , text=blue , minimum width =0.5 cm , minimum height =0.5 cm}]

\node[] (u1){};

\node[state, red, bottom color=pink] (bassGuitar)[left= of u1]{};
\node[] (bassGuitarText) [below= 0.5cm of bassGuitar]{Bass Guitar};

\node[state, red, bottom color=pink] (bass)[right= of u1]{};
\node[] (bassText) [below= 0.5cm of bass]{Bass};

\node[statebox] (t1)[above=of bassGuitar]{$t_1$};
\node[statebox] (t2)[above=of bass]{$t_2$};

\node[state, red, bottom color=pink] (will)[above= of t1]{};
\node[] (willText) [above= 0.5cm of will]{Will McGregor};

\node[state, red, bottom color=pink] (precious)[above= of t2]{};
\node[] (preciousText) [above= 0.5cm of precious]{Precious Things};

\path (will) edge [-] node [midway]  (anchor_1){}  (t1);
\node [left =0.35cm of anchor_1, rotate=90] {CONTRIB};

\path (precious) edge [-] node [midway]  (anchor_2){}  (t2);
\node [left =0.35cm of anchor_2, rotate=90] {RECORD};

\path (t1) edge [-] node [midway]  (anchor_3){}  (bassGuitar);
\node [left =0.35cm of anchor_3, rotate=90] {INSTR-ROLE};

\path (t2) edge [-] node [midway]  (anchor_4){}  (bass);
\node [left =0.35cm of anchor_4, rotate=90] {INSTR-ROLE};

%\path (t1) edge [-] node [midway]  (anchor_5){}  (u1);
%\node [right =0.1cm of anchor_5, rotate=-65] {FACT-ID};
%
%\path (t2) edge [-] node [midway]  (anchor_6){}  (u1);
%\node [above left=0.3cm and 0.3cm of anchor_6, rotate=65] {FACT-ID};

\path (will) edge [-] node [midway]  (anchor_7){}  (t2);
\node [above right =0.7cm and -0.6cm of anchor_7, rotate=-47] {CONTRIB};

\path (precious) edge [-] node [midway]  (anchor_8){}  (t1);
\node [below right = 0.65cm and -0.55cm of anchor_8, rotate=47] {RECORD};

\end{tikzpicture}
}

\end{tabular}
\end{center}
\caption{Fact $u_1$ in Table \ref{tab:fact2inst} (or Example \ref{ex:meta-relation}) converted to instances using $T_{\rm id}$ (left) and
$T$ (right) respectively. }
\label{fig:fact2inst}
\end{figure}
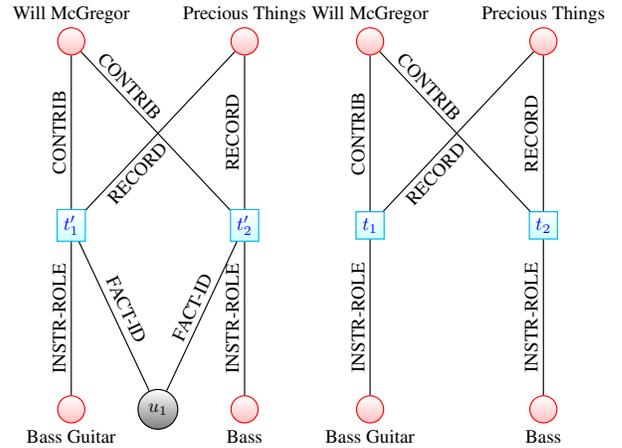

The reader is referred to Table \ref{tab:fact2inst} for an example converting a fact to instances using $T_{\rm id}(\cdot)$ and $T(\cdot)$.  Such conversions may also be viewed graphically as shown in Figure \ref{fig:fact2inst}.
Later in our experiments, we will investigate, relative to $T({\cal F})$, whether the fact-level information contained in $T_{\rm id}({\cal F})$ (namely, that certain instances belong to the same fact) is useful for embedding.

\begin{table*}[ht!]
\caption{\label{tab:fact2inst}
The facts $u_1$ and $u_2$ in Example \ref{ex:meta-relation} and conversions of $u_1$ to instances: $T(u_1): =\{t_1, t_2\}$, $T_{\rm id}(u_1): = \{t'_1, t'_2\}$.
}
\vspace{0.3cm}
\centerline{
\small
\begin{tabular}{|@{\hspace{0.1em}}
c@{\hspace{0.1em}}
|@{\hspace{0.1em}}
c@{\hspace{0.1em}}
|@{\hspace{0.1em}}
c@{\hspace{0.1em}}
|@{\hspace{0.1em}}
c@{\hspace{0.1em}}
|@{\hspace{0.1em}}
c@{\hspace{0.1em}}
|@{\hspace{0.1em}}
c@{\hspace{0.1em}}
|@{\hspace{0.1em}}
c@{\hspace{0.1em}}
|}
\hline
role & $u_1$(role) & $u_2$(role) &$t_1$(role) & $t_2$(role) & $t'_1$(role) & $t'_2$(role)  \\
\hline
\hline
${\rm CONTRIBUTOR}$ & $\{{\rm WillMcGregor}\}$  &  $\{{\rm MichaelHarrison}\}$& ${\rm WillMcGregor}$& ${\rm WillMcGregor}$ & ${\rm WillMcGregor}$ & ${\rm WillMcGregor}$ \\
\hline
${\rm RECORDING}$ & $\{{\rm PreciousThings}\}$  & $\{{\rm PrettyGoodYear}\}$& ${\rm PreciousThings}$ & ${\rm PreciousThings}$ &${\rm PreciousThings}$ & ${\rm PreciousThings}$  \\
\hline
${\rm INSTRUMENT\!\!\!-\!\!\!ROLE}$ & $\{{\rm BassGuitar}, {\rm Bass}\}$ & $\{{\rm Violin}\}$& ${\rm BassGuitar}$& ${\rm Bass}$& ${\rm BassGuitar}$& ${\rm Bass}$ \\
\hline
${\rm FACT\!\!\!-\!\!ID}$ &  &  &  & & $u_1$ & $u_1$ \\
\hline
\end{tabular}
}
%\vspace{-0.2cm}
\end{table*}

%\vspace{-0.2ex}

\subsection{Freebase}
\label{subsec:freebase}

Freebase \cite{FreebaseWeb} is a large collaborative KB. The core of Freebase may be understood as a fact representation,
containing about 3 billion facts and involving 50 million entities.  However,  following the RDF \cite{Google2015Review} format,  Freebase organizes its data as a collection of {\em triples}, which deviates from the fact representation defined in this paper. Let ${\cal F}$ denote the underlying fact representation of Freebase.  The essential format of Freebase can be obtained by manipulating ${\cal F}$ as follows.

For each fact in a binary degenerate meta-relation (defined in Section \ref{subsec:metaRelation}), apply the S2C conversion (Figure \ref{fig:star2clique}) to the fact vertex. This results in a collection of entity-predicate-entity triples, which are equivalent to instances of binary relations. The remaining fact vertices are called CVT  (Common Value Type) vertices in Freebase. Each labelled edge connected to a CVT vertex is represented as an entity-role-CVT triple. For each meta-relation $Q_q$ that is not binary degenerate, Freebase introduces a Mediator vertex to indicate the type $q$ of $Q_q$; then for each fact $u\in Q_q$, a Mediator-CONTAINS-CVT triple is created, where CVT indicates $u$ and Mediator indicates $q$, and CONTAINS is a fixed global token, independent of $u$ and $q$,  serving to mean $u\in Q_q$.

Using the three kinds of triples, Freebase's data organization
is equivalent to the fact representation ${\cal F}$. However, the heterogeneity in the triple semantics  makes this representation not as clean as the instance or fact representation, at least for embedding purpose.

\noindent
\begin{figure}[ht!]
%\vspace{-2ex}
\shadowbox{%
%\begin{minipage}{\dimexpr\textwidth-\shadowsize-2\fboxrule-2\fboxsep}
\begin{minipage}{0.94\linewidth}
   \textcolor{black}{\sffamily Star-to-Clique (S2C) Conversion}\par\vspace{\baselineskip}
%    \hspace*{\fill}
    %--------------------------%
 \vspace{-1ex}
On an edge labelled graph ${\cal G}$, let $(u, r, v)$ denote the labelled edge connecting vertices $u$ and $v$ with label $r$, and let ${\cal N}(v)$ denote the set of all adjacent vertices of $v$. Then the star-to-clique conversion on a vertex $s$ is defined by the following procedure.

\noindent (1) For every $x_1, x_2 \in {\cal N}(s)$ forming two labelled edges $(x_1, r_1, s)$ and  $(x_2, r_2, s)$ where $r_1\neq r_2$,  add a labelled edge  $(x_1, r_1.r_2, x_2)$ to ${\cal G}$.

\noindent (2) Delete $s$ and all edges connecting to $s$.

\vspace{1ex}

The reader is referred to Figure \ref{fig:s2c} for an example.
    \hspace*{\fill}
    %\par
   \vspace{0.2ex}
\end{minipage}}
%\vspace{-0.3cm}
\caption{\label{fig:star2clique} Definition of Star-to-Clique Conversion}
\end{figure}

\begin{figure}[ht!]
\begin{center}
\begin{tabular}{c@{\hspace{0.2cm}}c}

%\begin{figure}[ht!]
%\begin{center}
\scalebox{0.55}{
\begin {tikzpicture}[-latex ,auto ,node distance =1.5 cm and 1.5cm ,on grid ,
semithick ,
state/.style ={ circle ,top color =white , bottom color = processblue!20 ,
draw, processblue , text=blue , minimum width =0.5cm},
statebox/.style ={rectangle ,top color =white , bottom color = processblue!20 ,
draw, processblue , text=blue , minimum width =0.5 cm , minimum height =0.5 cm}]

\node[state, red, bottom color=pink] (e1){1};
\node[] (dum1)[right=of e1]{};
\node[state, red, bottom color=pink] (e2)[right=of dum1]  {2};
\node[statebox] (A)[below=of dum1]{$A$};
\node[] (dum2)[below=of A]  {};
\node[state, red, bottom color=pink] (e3)[left=of dum2]  {3};
\node[state, red, bottom color=pink] (e4)[right=of dum2]  {4};
\node[] (dum3)[right=of e4]  {};
\node[statebox] (B)[above=of dum3]{$B$};
\node[state, red, bottom color=pink] (e5)[right=of dum3]  {5};

\path (e1) edge [-] node [midway]  (anchor_e1A){}  (A);
\node [above right=0.04cm and 0.05cm of anchor_e1A] {$a$};

\path (e2) edge [-] node [midway]  (anchor_e2A){}  (A);
\node [above left =0.35cm and 0.2cm of anchor_e2A] {$b$};

\path (e3) edge [-] node [midway]  (anchor_e3A){}  (A);
\node [above left =0.00cm and 0.01cm of anchor_e3A] {$c$};

\path (e4) edge [-] node [midway]  (anchor_e4A){}  (A);
\node [above right =0.3cm and 0.3cm of anchor_e4A] {$d$};

\path (e4) edge [-] node [midway]  (anchor_e4B){}  (B);
\node [above left =0.01cm and 0.05cm of anchor_e4B] {$e$};

\path (e5) edge [-] node [midway]  (anchor_e5B){}  (B);
\node [above right =0.25cm and 0.4cm of anchor_e5B] {$f$};

\end{tikzpicture}
}
%\end{center}
%\caption{S2C before.}
%\label{fig:s2c1}
%\end{figure}

 &

%\begin{figure}[ht!]
%\begin{center}
\scalebox{0.55}{
\begin {tikzpicture}[-latex ,auto ,node distance =1.5 cm and 1.5cm ,on grid ,
semithick ,
state/.style ={ circle ,top color =white , bottom color = processblue!20 ,
draw, processblue , text=blue , minimum width =0.5cm},
statebox/.style ={rectangle ,top color =white , bottom color = processblue!20 ,
draw, processblue , text=blue , minimum width =0.5 cm , minimum height =0.5 cm}]

\node[state, red, bottom color=pink] (e1){1};
\node[] (dum1)[right=of e1]{};
\node[state, red, bottom color=pink] (e2)[right=of dum1]  {2};
\node[] (A)[below=of dum1]{};
\node[] (dum2)[below=of A]  {};
\node[state, red, bottom color=pink] (e3)[left=of dum2]  {3};
\node[state, red, bottom color=pink] (e4)[right=of dum2]  {4};
\node[] (dum3)[right=of e4]  {};
\node[] (B)[above=of dum3]{};
\node[state, red, bottom color=pink] (e5)[right=of dum3]  {5};

\path (e1) edge [-] node [midway]  (e12){}  (e2);
\node [above =0.1cm of e12] {$a.b$};

\path (e3) edge [-] node [midway]  (e34){}  (e4);
\node [above =0.1cm of e34] {$c.d$};

\path (e4) edge [-] node [midway]  (e45){}  (e5);
\node [above =0.1cm of e45] {$e.f$};

\path (e1) edge [-] node [midway]  (e13){}  (e3);
\node [left =0.5cm of e13] {$a.c$};

\path (e2) edge [-] node [midway]  (e24){}  (e4);
\node [right =0.3cm of e24] {$b.d$};

\path (e1) edge [-] node [midway]  (e14){}  (e4);
\node [above left =0.2cm and 0.8cm of e14] {$a.d$};

\path (e2) edge [-] node [midway]  (e23){}  (e3);
\node [above right =0.4cm and 0.6cm of e23] {$b.c$};

\end{tikzpicture}
}
%\end{center}
%\caption{S2C before.}
%\label{fig:s2c1}
%\end{figure}

\end{tabular}
\end{center}
\caption{Applying the S2C conversion to vertices $A$ and $B$ in the labelled graph on the left results in the labelled graph on the right.}
\label{fig:s2c}
\end{figure}
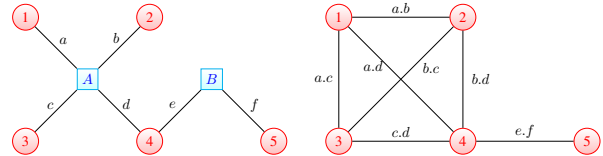

%\vspace{-0.3cm}
\section{Embedding}
%\vspace{-0.4ex}
\subsection{Problem Formulation}
%\vspace{-0.3ex}

Instance representations, fact representations, and the RDF-based triple representations (e.g., that in Freebase) may all contain equivalent structural information about the relations contained in a KB. However, instance representations have a uniform semantics comparing with the RDF-based triple representations and are easier to deal with than fact representations. For this reason, we now formulate the KB embedding problem on an instance representation $\left({\cal N}, {\cal R}, \{{\cal T}_r: r\in {\cal R}\}\right)$.

Let a vector space $U$ over the field ${\mathbb R}$ of real numbers
be the chosen space for embedding. The objective of KB embedding is to  construct a function $\phi: {\cal N} \rightarrow U$ and a subset $C_r\subset U^{{\cal M}(R_r)}$ for each relation $R_r$ such that  {\em ideally} the following properties are satisfied.
 \begin{enumerate}
 \item For every $r\in {\cal R}$ and every instance $t\in {R}_r$,  $\phi\circ t \in C_r$, where the symbol $\circ$ denotes function composition \cite{hungerford}.
\item For every $r\in {\cal R}$ and every function $t\in {\cal N}^{{\cal M}(R_r)}\setminus{R}_r$,  $\phi\circ t \notin C_r$.
 \end{enumerate}
Here, the function $\phi$, serving as a representation of ${\cal N}$,  maps an entity to its embedding vector.  The subsets $\{C_r: r\in {\cal R}\}$,   serving  as a representation of $\{R_r: r\in {\cal R}\}$, define a set of  constraints on the embedding vectors which preserve the intra-relational and inter-relational structures of $\{R_r: r\in {\cal R}\}$.

 Note that each constraint $C_r$ may be identified  with a non-negative cost function $f_r:U^{{\cal M}(R_r)}\rightarrow {\mathbb R}$ such that
 \begin{eqnarray}
 \label{eq:sat}
 f_r({\bf t}) = 0 & {\rm if} & {\bf t}\in C_r, ~{\rm and} \\
 \label{eq:unsat}
  f_r({\bf t}) > 0 & {\rm if} & {\bf t}\notin C_r
 \end{eqnarray}
 Denote $\Theta:= \{f_r: r\in {\cal R}\}$. The problem then translates to determining $(\Theta, \phi)$. But
 $\{R_r: r\in {\cal R}\}$ is unknown, and all we have is the observed instances $\{{\cal T}_r: r\in {\cal R}\}$
 and possibly some ``negative examples'' $\{{\cal T}^-_r: r\in {\cal R}\}$, where each ${\cal T}^-_r \subset
 {\cal N}^{{\cal M}(R_r)}\setminus R_r$. Note that when the KB is large, for any $t\in {\cal T}_r$, if we replace its value $t(\rho)$ for some role $\rho \in {\cal M}(R_r)$ with a random entity,  the resulting function falls in  ${\cal N}^{{\cal M}(R_r)}\setminus R_r$ with high probability. This can be used to construct ${\cal T}^-_r$ (\cite{transE,transH,transR}).

Treating the problem as {\em learning} $(\Theta, \phi)$, we may not need the property 1 above to hold strictly. Then the equality ``$=0$"  in (\ref{eq:sat}) is taken as ``as close to $0$ as possible''. Towards a margin-based optimization formulation (which gives better discriminative power and robustness), the threshold $0$ in (\ref{eq:unsat}) is raised to a positive value $c$. The problem can then be formulated as finding $({\Theta}, \phi)$ to minimize the following global cost function.
 \begin{equation}
\label{eq:obj}
F({\Theta}, \phi)\!\!:=\!\!\!\sum\limits_{r\in {\cal R}}\!\!
\left(
\sum\limits_{t\in {\cal T}_r} f_{r}(\phi\circ t)\!\!
+ \!\!\!\!\!\!\sum\limits_{t^-\in {\cal T}^{-}_{r}} \!\!\!\!\left[c - f_{r}(\phi\circ t^-)\right]_{+}
\right),
\end{equation}
where $[\cdot]_+$ denotes the rectifier function \cite{rectifierNetwork}, namely, $[a]_{+}: = \max(0, a)$.

What remains is to choose a proper space of $\Theta$ for this  optimization problem, which is at the heart of modelling.

%\vspace{-0.3ex}

\subsection{FB15K Datasets}
%\vspace{-0.3ex}

A popular dataset for training and testing embedding models is the FB15K datasets \cite{transE}, filtered from Freebase.  When viewing Freebase as an edge-labelled graph (i.e., every triple as an labelled edge), we observe that to every CVT vertex in the filtered data,  FB15K has applied the S2C conversion (Figure \ref{fig:star2clique}).
\vspace{-0.1cm}

\begin{lem}
\label{lem:s2c}
After applying S2C conversions to a graph ${\cal G}$,  in general, ${\cal G}$ is no longer recoverable.
\end{lem}
This suggests a loss of structural information in this conversion.
Working with this dataset, for which the original fact or instance representation is no longer recoverable,  one is left with no option for embedding multi-fold relations but to treat the triples as instances of binary relations. In addition, we observe that the S2C conversions applied in FB15K have also involved the Mediator vertices connected to the CVTs. This results in
each Mediator vertex in FB15K connecting to a good number of entity vertices. This connectivity provides no information about the structures of the relations, arguably only serving as ``noise'' for embedding.  For these reasons and since we want to work with multi-fold relational data in their intact form,  FB15K no longer suits our purpose.

%\vspace{-0.3ex}

\subsection{Prior Art of Modelling}
%\vspace{-0.3ex}

To date, well-known models developed for KB embedding include TransE \cite{transE}, {TransH}\cite{transH}, {TransR} \cite{transR}, the {Unstructured Model} (UE) \cite{UE_SME}, the {Structure Embedding Model} \cite{SE}, the { Neural Tensor Network}  model\cite{SLM} and the {Single Layer Model}\cite{SLM}.
All these models deal with datasets in a triple representation (such as FB15K) and treat each triple as an instance of a binary relation. This treatment is equivalent to regarding the triple  representation as an instance representation where each relation $R_r$ is binary. Consequently, each cost function $f_r$ may be regarded as a function on $U^2$.  For example, in TransH, the function $f_r$ is defined by
\begin{equation}
\label{eq:transH_F}
f_r({\bf x}, {\bf y}) = \lVert {\mathbb P}_{{\bf n}_r}({\bf x})+ {\bf d}_r - {\mathbb P}_{{\bf n}_r}({\bf y})
\rVert^2,
\end{equation}
where ${\bf n}_r$ a unit-length vector in $U$, ${\bf d}_r$ is a vector in the hyperplane in $U$ with normal vector ${\bf n}_r$, and ${\mathbb P}_{{\bf n}_r}: U\rightarrow U$ is the function that maps a ${\bf z}\in U$ to the projection of ${\bf z}$ on the hyperplane with normal vector ${\bf n}_r$, namely,
\[
{\mathbb P}_{{\bf n}_r}({\bf z}): = {\bf z} - {\bf z}^T{\bf n}_r{\bf n}_r.
\]

Among these models, TransE is arguably the most influential, which has inspired the later models in the ``Trans series". TransR is reportedly the best performing model \cite{transR}.  TransH slightly under-performs TransR on FB15K, but having much lower computation complexity.

\noindent{\em Decomposition Framework}:  Let ${\cal G}:=({\cal N}, {\cal R}, \{{\cal T}_r: r\in {\cal R}\})$ be an instance representation.  Let $\Gamma_r$ be the set of all size-2 subsets of ${\cal M}(R_r)$.   In the decomposition framework, every $f_r: U^{{\cal M}(R_r)}\rightarrow {\mathbb R}$ is parametrized as
\begin{equation}
\label{eq:decomp}
f_r({\bf t}):=\sum\limits_{\gamma\in \Gamma_r} f_{\gamma}({\bf t}_{:\gamma}).
\end{equation}
Let ${\rm S2C}({\cal G})$ denote the triple representation resulting  from S2C-converting every instance vertex in ${\cal G}$.

%\vspace{-0.1cm}

\begin{lem} The global cost function (\ref{eq:obj}) for ${\rm S2C}({\cal G})$ is equivalent to
the global function (\ref{eq:obj}) for ${\cal G}$ under the parametrization
(\ref{eq:decomp}).
\end{lem}
That is,  models on a triple representation resulting from S2C conversion of an instance representation are equivalent to modeling the original instance representation using the decomposition framework. Noting that the FB15K data have undergone the S2C conversion, all afore-mentioned models on FB15K are, coincidentally, special cases of the decomposition framework.
Since the S2C conversion distorts the structures of the relations in the KB
(Lemma \ref{lem:s2c}), the decomposition framework necessarily suffers from a loss of information.

In addition, it is possible to show that parametrization using the decomposition framework can result in large errors in approximating the function $f_r$.

Recently another model PTransE\cite{PtransE} has been proposed. An extension of TransE without considering multi-fold relations, PTransE  performs comparably to TransR.

%\vspace{-0.3ex}

\subsection{Proposed Model}
%\vspace{-0.3ex}

As the decomposition framework is fundamentally limited,  we advocate a {\em direct modelling framework}, namely, that the cost function $f_r$ is modelled directly without recourse to the decomposition in (\ref{eq:decomp}).  We now present an example model in this framework, termed {\em m-TransH}, which generalizes TransH directly to multi-fold relations.

In {\em m-TransH}, each cost function $f_r$ is parametrized by two unit-length orthogonal vectors ${\bf n}_r$ and ${\bf b}_r$ in $U$ and
a function ${\bf a}_r \in {\mathbb R}^{{\cal M}(R_r)}$. More specifically, the  function $f_r$ is defined by
\begin{equation}
\label{eq:mTransH}
f_r({\bf t}) : = \left\lVert
\sum\limits_{\rho\in {\cal M}(R_r)}
{\bf a}_r(\rho) {\mathbb P}_{{\bf n}_r}({\bf t}(\rho))
+ {\bf b}_r \right\rVert^2, ~{\bf t}\in {\cal N}^{{\cal M}(R_r)}.
\end{equation}

In addition, the orthogonality and unit-lengths constraints on the parameters ${\bf b}_r$ and
${\bf n}_r$ is implemented as L2 penalizing terms added to the cost function $f_r$ in (\ref{eq:mTransH}).

\begin{lem} If each $R_r$ is a binary relation and
$\sum_{\rho\in {\cal M}(R_r)} {\bf a}_r(\rho) = 0$,  the optimization problem of m-TransH (specified via (\ref{eq:obj}) and (\ref{eq:mTransH})) and that of TransH (specified via (\ref{eq:obj}) and (\ref{eq:transH_F})) are identical.
\end{lem}
Thus m-TransH reduces to TransH for binary relations.

%\vspace{-0.4cm}
\section{Experiments}

%\vspace{-0.2ex}

\subsection{JF17K Datasets}
%\vspace{-0.2ex}

The full Freebase data in RDF format was downloaded.  Entities involved in very few triples and the triples involving {\em String}, {\em Enumeration Type} and {\em Numbers} were removed. A fact representation was recovered from the remaining triples.  Facts from meta-relations having only a single role
% (e.g., the meta-relation representing the sibling relationship)
were removed.  From each meta-relation containing more than 10000 facts, 10000 facts were randomly selected. Denote the resulting fact representation by ${\cal F}$. Two instance representations\footnote{In the construction of $T_{\rm id}({\cal F})$, when acting on facts of degenerate meta-relations, $T_{\rm id}$ is taken as $T$. Such facts each contain a single instance, and there is no need to keep their ID's.} $T_{\rm id}({\cal F})$ and $T({\cal F})$
were constructed. Further filtering was applied to  $T({\cal F})$ such that each entity is involved in at least 5 instances.
Denote the filtered  $T({\cal F})$  by ${\cal G}$.
 Then $T_{\rm id}({\cal F})$ was filtered correspondingly so that it contains the same set of instances as ${\cal G}$. Denote the filtered $T_{\rm id}({\cal F})$ by  ${\cal G}_{\rm id}$.  Instance representation ${\rm S2C}({\cal G})$ was constructed and is  denoted  by ${\cal G}_{\rm s2c}$. This resulted in three consistent datasets, ${\cal G}$, ${\cal G}_{\rm id}$ and ${\cal G}_{\rm s2c}$.

Dataset ${\cal G}_{\rm id}$ was randomly split into training set ${\cal G}_{\rm id}^{\checkmark}$ and testing set ${\cal G}_{\rm id}^{?}$ where every fact ID  entity in ${\cal G}_{\rm id}^{?}$ was assured to appear in ${\cal G}_{\rm id}^{\checkmark}$. The corresponding splitting was then applied to
${\cal G}$ and ${\cal G}_{\rm s2c}$, giving rise to training sets
${\cal G}^{\checkmark}$ and ${\cal G}_{\rm s2c}^{\checkmark}$, and testing sets ${\cal G}^{?}$ and ${\cal G}_{\rm s2c}^{?}$. We call these datasets JF17K. Their statistics, in the same order as FB15K,
are given in Table~\ref{tab:dataStats}.
%in Table \ref{tab:dataStats}.
\vspace{-0.4cm}
\begin{table}[!h]
\caption{\label{tab:dataStats} Statistics of JF17K.}
\vspace{0.3cm}
%\begin{table}
\centerline{\small
\begin{tabular}{|@{\hspace{0.3em}}
l@{\hspace{0.3em}}
|@{\hspace{0.3em}}
c@{\hspace{0.3em}}
|@{\hspace{0.3em}}
c@{\hspace{0.3em}}
|@{\hspace{0.3em}}
c@{\hspace{0.3em}}
|@{\hspace{0.3em}}
c@{\hspace{0.3em}}
|}
\hline
& ${\cal G}^{\checkmark}$/${\cal G}_{\rm id}^{\checkmark}$ & ${\cal G}_{\rm s2c}^{\checkmark}$ & ${\cal G}^{?}$/${\cal G}_{\rm id}^{?}$ & ${\cal G}_{\rm s2c}^{?}$\\
\hline
\hline
\# of entities & 17629 & 17629  & 12282 & 12282 \\
\hline
\# of instances/triple types & 181 &381 & 159 & 336 \\
\hline
\# of instances/triples & 139997 & 254366 & 22076 & 52933 \\
\hline
\end{tabular}
}
\end{table}

%\vspace{-0.2ex}

\subsection{Training and Testing}
%\vspace{-0.2ex}

We performed four kinds of experiments,  termed {\tt m-TransH},  {\tt m-TransH:ID}, {\tt TransH:triple}  and {\tt TransH:inst}, in which {\tt m-TransH} and {\tt m-TransH:ID} train an  m-TransH model,  and {\tt TransH:triple}  and {\tt TransH:inst} train a
TransH model. The training and testing datasets for the experiments are given in Table~\ref{tab:TTD}.
\vspace{-0.4cm}
\begin{table}[!h]
\caption{\label{tab:TTD}The training and testing datasets for the experiments.}
\vspace{.3cm}
\centerline{
{\small
\begin{tabular}{
|@{\hspace{0.5em}}
c@{\hspace{0.5em}}
|@{\hspace{0.5em}}
c@{\hspace{0.5em}}
|@{\hspace{0.5em}}
c@{\hspace{0.5em}}
|@{\hspace{0.5em}}
c@{\hspace{0.5em}}
|}
\hline
 {  {\fontsize{10.5pt}{18pt}\tt} m-TransH} & {  {\fontsize{10.5pt}{18pt}\tt} m-TransH:ID} & {  {\fontsize{10.5pt}{18pt}\tt} TransH:triple} & {  {\fontsize{10.5pt}{18pt}\tt} TransH:inst} \\
\hline
\hline
 $({\cal G}^{\checkmark}, {\cal G}^?)$ & $({\cal G}_{\rm id}^{\checkmark}, {\cal G}_{\rm id}^?)$ &$({\cal G}_{\rm s2c}^{\checkmark}, {\cal G}_{\rm s2c}^?)$ & $({\cal G}_{\rm s2c}^{\checkmark}, {\cal G}_{\rm s2c}^?)$\\
\hline
\end{tabular}
}
}
\end{table}

\vspace{1ex}

\noindent In {\tt TransH:triple}  and {\tt TransH:inst}, for each triple in ${\cal G}_{\rm sc}^{\checkmark}$, one random negative example is generated. In {\tt m-TransH} and {\tt m-TransH:ID}, for each instance in ${\cal G}^{\checkmark}$,
%(or in ${\cal G}_{\rm id}^{\checkmark}$),
${L \choose 2}$ random negative examples are generated. This way, the total number of negative examples used in every experiment is the same, assuring a fair comparison. Stochastic Gradient Descent is used for training, as is standard. Several choices of the dimension (DIM) of $U$ are studied.

\noindent{\em Testing Protocol}:
In {\tt m-TransH}, {\tt m-TransH:ID} and {\tt TransH:triple}, for each testing instance/triple and for each entity $x$ therein,  assume that $x$ is unknown; evaluate the cost function for the embedding of the instance/triple by replacing $x$ with every $x'\in {\cal N}$, rank the cost of $x'$ from low to high, and record the rank for $x'=x$. Hit@10 (HIT) and Mean Rank (RANK) are used as the performance metrics\cite{transE}. Since the number of triples in ${\cal G}^{?}_{\rm s2c}$ is significantly larger than the number of instances in ${\cal G}^{?}_{\rm id}$ and ${\cal G}^{?}$, it is questionable whether this testing protocol is fair to the TransH model used in {\tt TransH:triple}. To assure fairness, in {\tt TransH:inst},  the TransH model is interpreted as parametrizing the cost function for each relation type in ${\cal G}^{?}$ using the decomposition framework (\ref{eq:decomp}); then for each instance in ${\cal G}^{?}$, each entity $x$ therein is queried only once in {\tt TransH:inst}, and the cost for each replacing entity is computed at the instance level instead,  using (\ref{eq:decomp}) and (\ref{eq:transH_F}).

%\vspace{-0.2ex}

\subsection{Results and Discussions}

%\vspace{-0.2ex}

\begin{figure}[!htb]
  \centering
 % \includegraphics[width=9cm]{dim}\\
   \includegraphics[width=1.0\columnwidth]{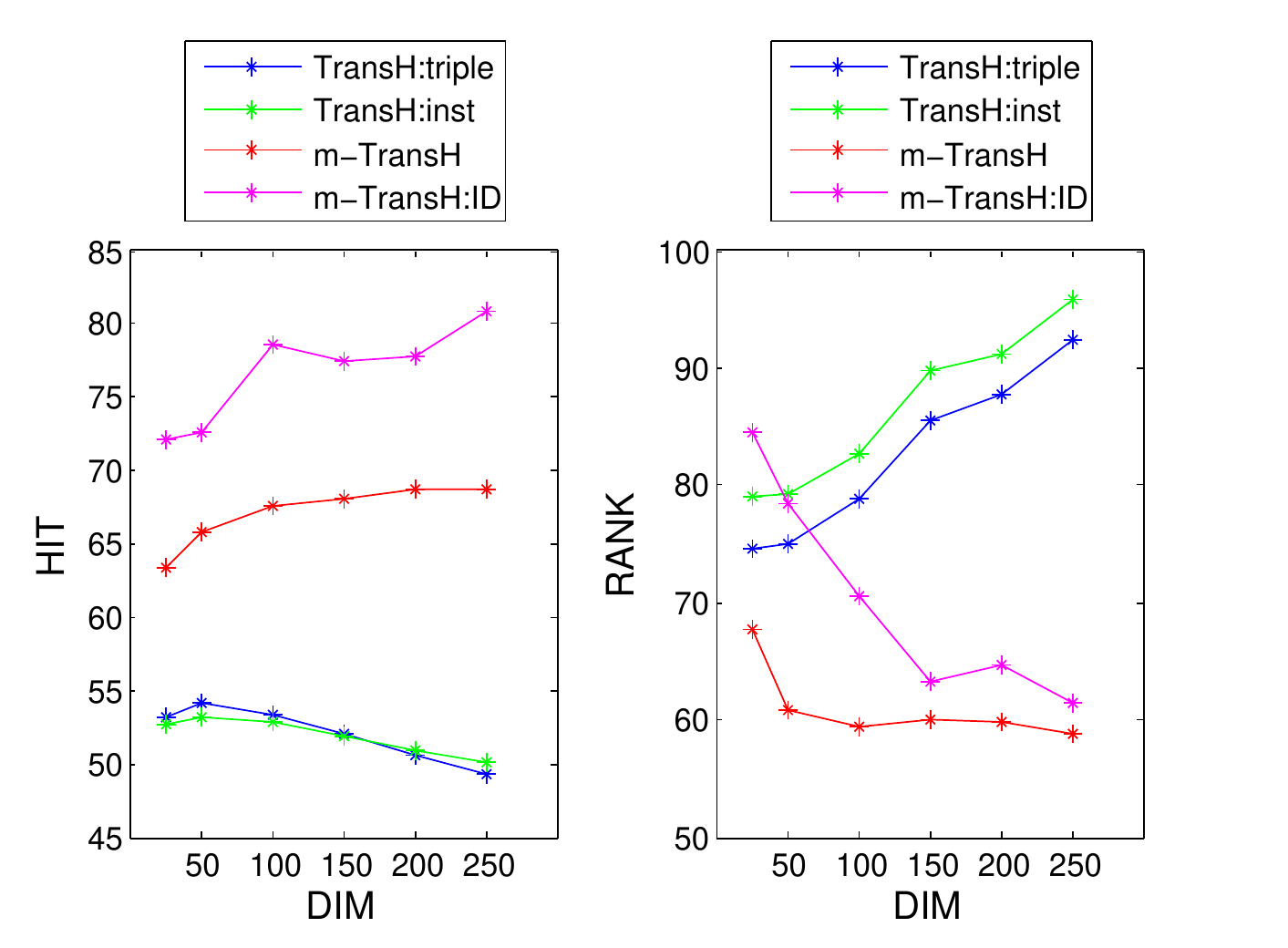}\\
  %\vspace{-0.5cm}
  \caption{Overall  performances in Table \ref{tab:overall}.}
  \label{fig:overall}
\end{figure}

The overall performances of the four kinds of experiments are shown in Table \ref{tab:overall} and Figure \ref{fig:overall}.

With the two different testing protocols, TransH performs similarly, where the achieved HIT values are virtually identical and the instance-level test appears somewhat inferior in RANK. This suggests that using the triple-level test, {\tt TransH:triple},  to compare with m-TransH at least won't depreciate TransH. Comparing with its reported performance on FB15K \cite{transH}, TransH performs  rather significantly better on JF17K. This, to an extent attributed to the intrinsic difference in the data,  may also reflect that the filtering process of JF17K is more delicate. For example, JF17K contains no Mediator vertices, hence ``cleaner''.

Overall m-TransH on both ${\cal G}^{?}$ and ${\cal G}^{?}_{\rm id}$ outperforms TransH by a huge margin. We believe that this performance gain is solely due to the fact that multi-fold relations are treated properly with the direct modelling framework. It is pleasant to see that HIT above 80\% can be achieved in this framework, significantly exceeding all reported embedding performances (which are about 50\%).  This leap perhaps signals that KB embedding is at the verge of prevailing in practice.  The observation that {\tt m-TransH:ID} consistently outperforms {\tt m-TransH} in HIT implies that fact-level information can be useful for embedding, a direction certainly deserving further investigation. However, the RANK positions of two m-TransH settings are reversed: {\tt m-TransH:ID} under-performs {\tt m-TransH}, and the RANK gap decreases as DIM increases. This is because a large number (36201) of fact IDs, twice the number of entities, need to be embedded in {\tt m-TransH:ID}. When DIM is low, there are not sufficient degrees of freedom to fit all the entities and the fact IDs. This makes a fraction of entities embedded particularly poorly and brings the overall RANK worse than {\tt m-TransH}. As DIM increases, this problem becomes less severe, resulting in improved RANK.

For each kind of experiments, we select the dimension (DIM) at which it performs the best and break down its performance across relations with different fold $J$. We observe in Table \ref{tab:breakdown} that
%  and Figure \ref{fig:breakdown},
for {\em every} fold value $J$,  m-TransH outperforms TransH. Note that m-TransH is designed to handle multi-fold relations. However its improved embedding performance for non-binary relations clearly also has a global impact,  allowing the binary relations to be embedded better as well.

Finally, we note that the time complexity of m-TransH is significantly lower than TransH.
For example, at DIM=50, the training/testing times (in minutes) for {\tt TransH:triple} and {\tt m-TransH:ID} are respectively 105/229 and 52/135, on a 32-core Intel E5-2650 2.0GHz processor.
This is because with the decomposition framework, an instance of a $J$-fold relation is S2C-converted to ${J \choose 2}$ triples, greatly increasing the number of model parameters.

\begin{table}
\caption{\label{tab:overall} Overall performances (HIT/RANK)}
\vspace{0.3cm}
\centerline{\small
\begin{tabular}{|@{\hspace{0.1em}}c@{\hspace{0.1em}}|@{\hspace{0.1em}}c@{\hspace{0.1em}}|@{\hspace{0.1em}}c@{\hspace{0.1em}}|@{\hspace{0.1em}}c@{\hspace{0.1em}}|@{\hspace{0.1em}}c@{\hspace{0.1em}}|}
\hline
DIM  & {{\fontsize{10.5pt}{18pt}\tt} TransH:triple} & {{\fontsize{10.5pt}{18pt}\tt} TransH:inst} & {{\fontsize{10.5pt}{18pt}\tt} m-TransH} & {{\fontsize{10.5pt}{18pt}\tt} m-TransH:ID}\\
\hline
\hline
25&  53.12\%/{\bf 74.6} &52.64\%/{\bf 79.0} & 63.45\%/67.8 &72.12\%/84.5\\
\hline
50&  {\bf 54.21}\%/75.0 &{\bf 53.10}\%/79.2 & 65.87\%/60.8 &72.54\%/78.5\\
\hline
100& 53.32\%/78.9 &52.86\%/82.7 & 67.54\%/59.4 &78.61\%/70.5\\
\hline
150& 52.11\%/85.4 &51.91\%/89.7 & 68.12\%/60.0 &77.51\%/63.3\\
\hline
200& 50.54\%/87.8 &50.98\%/91.1 & {\bf 68.75}\%/59.8 &77.81\%/64.7\\
\hline
250& 49.26\%/92.4 &50.10\%/95.9 & 68.73\%/{\bf 58.7} & {\bf 80.82}\%/{\bf 61.4}\\
\hline
\end{tabular}
}
%\vspace{-0.3cm}
\end{table}

\begin{table}
\caption{\label{tab:breakdown} Breakdown performances (HIT/RANK) across relations with different fold $J$.}
\vspace{0.3cm}
\centerline{\small
\begin{tabular}{|@{\hspace{0.3em}}c@{\hspace{0.3em}}|@{\hspace{0.3em}}c@{\hspace{0.3em}}|@{\hspace{0.3em}}c@{\hspace{0.3em}}|@{\hspace{0.3em}}c@{\hspace{0.3em}}|@{\hspace{0.3em}}c@{\hspace{0.3em}}|}
\hline
 & {{\fontsize{10.5pt}{18pt}\tt} TransH:triple} & {{\fontsize{10.5pt}{18pt}\tt} TransH:inst} & {{\fontsize{10.5pt}{18pt}\tt} m-TransH} & {{\fontsize{10.5pt}{18pt}\tt} m-TransH:ID}\\[-0.7ex]
\raisebox{1ex}{$J$}& DIM=50 & DIM=50 & DIM=250 & DIM=250\\
\hline
\hline
2 & 50.29\%/102.7 & 50.30\%/102.4  & 58.60\%/78.7 & 65.96\%/99.3 \\
\hline
3 & 49.99\%/86.6 & 50.79\%/77.5 & 67.73\%/60.3 & 90.38\%/39.2 \\
\hline
4 & 56.58\%/56.2 & 59.69\%/43.1  & 90.19\%/18.5 & 92.10\%/32.2 \\
\hline
5 & 75.93\%/8.1 & 76.13\%/7.0  & 93.93\%/4.7 & 95.36\%/5.8 \\
\hline
6 & 98.52\%/7.8 & 100\%/1.5  & 100\%/1.5 & 100\%/2.0 \\
\hline
%7 & 00.00\%/00.00 & 00.00\%/00.00  & 00.00\%/00.00 &00.00\%/00.00 \\
%\hline
\end{tabular}
}
%\vspace{-0.3cm}
\end{table}

%\vspace{-0.5cm}

\section{Concluding Remarks}

%\vspace{-0.1cm}

This paper examines the fundamentals of multi-fold relations and advocates instance representations as a canonical representation for knowledge bases. We show that the widely adopted decomposition framework,  which models multi-fold relations at the triple level,  is fundamentally limited. Instead,  we propose to model multi-fold relations at the instance level. With a simple example of such models, we demonstrate great advantages of this approach both in performance and in complexity. Outperforming TransH by a cheerful margin, this simple model, m-TransH, perhaps signals the arrival of a new performance regime in knowledge base embedding.

To inspire further research on the embedding of multi-fold relations,  we have made our JF17K datasets publicly available.\footnote{\url{http://www.site.uottawa.ca/~yymao/JF17K}}

\section{Acknowledgement}
This work is supported partly by China 973 program (No. 2014CB340305), partly by the National Natural Science Foundation of China (No. 61300070, 61421003), and partly by the Beijing Advanced Innovation Center for Big Data and Brain Computing.

%\clearpage

%\bibliographystyle{abbrv}
\bibliographystyle{named}

\begin{thebibliography}{}

\end{thebibliography}


\begin{thebibliography}{}

\bibitem[\protect\citeauthoryear{Angeli and Manning}{2013}]{kgCompleting2013}
Gabor Angeli and Christopher~D Manning.
\newblock Philosophers are mortal: Inferring the truth of unseen facts.
\newblock In {\em CoNLL}, pages 133--142, 2013.

\bibitem[\protect\citeauthoryear{Auer \bgroup \em et al.\egroup
  }{2007}]{auer2007dbpedia}
S{\"o}ren Auer, Christian Bizer, Georgi Kobilarov, Jens Lehmann, Richard
  Cyganiak, and Zachary Ives.
\newblock {\em Dbpedia: A nucleus for a web of open data}.
\newblock Springer, 2007.

\bibitem[\protect\citeauthoryear{Baader \bgroup \em et al.\egroup
  }{2007}]{kgCompleting2007}
Franz Baader, Bernhard Ganter, Baris Sertkaya, and Ulrike Sattler.
\newblock Completing description logic knowledge bases using formal concept
  analysis.
\newblock In {\em IJCAI}, volume~7, pages 230--235, 2007.

\bibitem[\protect\citeauthoryear{Bollacker \bgroup \em et al.\egroup
  }{2008}]{bollacker2008freebase}
Kurt Bollacker, Colin Evans, Praveen Paritosh, Tim Sturge, and Jamie Taylor.
\newblock Freebase: a collaboratively created graph database for structuring
  human knowledge.
\newblock In {\em Proceedings of the 2008 ACM SIGMOD international conference
  on Management of data}, pages 1247--1250. ACM, 2008.

\bibitem[\protect\citeauthoryear{Bordes \bgroup \em et al.\egroup }{2011}]{SE}
Antoine Bordes, Jason Weston, Ronan Collobert, and Yoshua Bengio.
\newblock Learning structured embeddings of knowledge bases.
\newblock In {\em Conference on Artificial Intelligence}, number
  EPFL-CONF-192344, 2011.

\bibitem[\protect\citeauthoryear{Bordes \bgroup \em et al.\egroup
  }{2013}]{transE}
Antoine Bordes, Nicolas Usunier, Alberto Garcia-Duran, Jason Weston, and Oksana
  Yakhnenko.
\newblock Translating embeddings for modeling multi-relational data.
\newblock In {\em Advances in Neural Information Processing Systems}, pages
  2787--2795, 2013.

\bibitem[\protect\citeauthoryear{Bordes \bgroup \em et al.\egroup
  }{2014a}]{qasubg2014}
Antoine Bordes, Sumit Chopra, and Jason Weston.
\newblock Question answering with subgraph embeddings.
\newblock {\em arXiv preprint arXiv:1406.3676}, 2014.

\bibitem[\protect\citeauthoryear{Bordes \bgroup \em et al.\egroup
  }{2014b}]{UE_SME}
Antoine Bordes, Xavier Glorot, Jason Weston, and Yoshua Bengio.
\newblock A semantic matching energy function for learning with
  multi-relational data.
\newblock {\em Machine Learning}, 94(2):233--259, 2014.

\bibitem[\protect\citeauthoryear{Codd}{1970}]{nAryDB}
Edgar~F Codd.
\newblock A relational model of data for large shared data banks.
\newblock {\em Communications of the ACM}, 13(6):377--387, 1970.

\bibitem[\protect\citeauthoryear{Fre}{2016}]{FreebaseWeb}
{Freebase}.
\newblock \url{http://www.freebase.com}, 2016.
\newblock Accessed: 2016-01-20.

\bibitem[\protect\citeauthoryear{Glorot \bgroup \em et al.\egroup
  }{2011}]{rectifierNetwork}
Xavier Glorot, Antoine Bordes, and Yoshua Bengio.
\newblock Deep sparse rectifier neural networks.
\newblock In {\em International Conference on Artificial Intelligence and
  Statistics}, pages 315--323, 2011.

\bibitem[\protect\citeauthoryear{Grewe}{2010}]{NARY-IMP2}
N~Grewe.
\newblock A generic reification strategy for n-ary relations in dl.
\newblock In H.~Herre et~al., editors, {\em Proceedings of the 2nd workshop of
  the GI-Fachgruppe 'Ontologien in Biomedizin und Lebenswissenschaften' (OBML)
  : Mannheim, Germany}, 2010.

\bibitem[\protect\citeauthoryear{Hungerford}{2003}]{hungerford}
Thomas Hungerford.
\newblock {\em Algebra}.
\newblock Springer, eighth edition, 2003.

\bibitem[\protect\citeauthoryear{Krieger and Willms}{2015}]{NARY-IMP3}
Hans-Ulrich Krieger and Christian Willms.
\newblock Extending owl ontologies by cartesian types to represent n-ary
  relations in natural language.
\newblock In {\em Language and Ontologies 2015. Language and Ontologies,
  located at 11th International Conference on Computational Semantics, April
  14, London, United Kingdom}. o.A., 4 2015.

\bibitem[\protect\citeauthoryear{Lin \bgroup \em et al.\egroup
  }{2015a}]{PtransE}
Yankai Lin, Zhiyuan Liu, and Maosong Sun.
\newblock Modeling relation paths for representation learning of knowledge
  bases.
\newblock {\em arXiv preprint arXiv:1506.00379}, 2015.

\bibitem[\protect\citeauthoryear{Lin \bgroup \em et al.\egroup
  }{2015b}]{transR}
Yankai Lin, Zhiyuan Liu, Maosong Sun, Yang Liu, and Xuan Zhu.
\newblock Learning entity and relation embeddings for knowledge graph
  completion.
\newblock In {\em Proceedings of AAAI}, 2015.

\bibitem[\protect\citeauthoryear{Marin \bgroup \em et al.\egroup
  }{2014}]{KG-APP1}
Alex Marin, Roman Holenstein, Ruhi Sarikaya, and Mari Ostendorf.
\newblock Learning phrase patterns for text classification using a knowledge
  graph and unlabeled data.
\newblock In {\em Fifteenth Annual Conference of the International Speech
  Communication Association}, 2014.

\bibitem[\protect\citeauthoryear{Nguyen \bgroup \em et al.\egroup
  }{2014}]{KGNARY-1}
Vinh Nguyen, Olivier Bodenreider, and Amit Sheth.
\newblock Don't like rdf reification?: Making statements about statements using
  singleton property.
\newblock In {\em Proceedings of the 23rd International Conference on World
  Wide Web}, WWW '14, pages 759--770, New York, NY, USA, 2014. ACM.

\bibitem[\protect\citeauthoryear{Nickel \bgroup \em et al.\egroup
  }{2015}]{Google2015Review}
Maximilian Nickel, Kevin Murphy, Volker Tresp, and Evgeniy Gabrilovich.
\newblock A review of relational machine learning for knowledge graphs: From
  multi-relational link prediction to automated knowledge graph construction.
\newblock {\em arXiv preprint arXiv:1503.00759}, 2015.

\bibitem[\protect\citeauthoryear{Rouces \bgroup \em et al.\egroup
  }{2015}]{NARY-IMP1}
Jacobo Rouces, Gerard de~Melo, and Katja Hose.
\newblock Framebase: Representing n-ary relations using semantic frames.
\newblock In {\em The Semantic Web. Latest Advances and New Domains}, pages
  505--521. Springer, 2015.

\bibitem[\protect\citeauthoryear{Socher \bgroup \em et al.\egroup }{2013}]{SLM}
Richard Socher, Danqi Chen, Christopher~D Manning, and Andrew Ng.
\newblock Reasoning with neural tensor networks for knowledge base completion.
\newblock In {\em Advances in Neural Information Processing Systems}, pages
  926--934, 2013.

\bibitem[\protect\citeauthoryear{Suchanek \bgroup \em et al.\egroup
  }{2007}]{suchanek2007yago}
Fabian~M Suchanek, Gjergji Kasneci, and Gerhard Weikum.
\newblock Yago: a core of semantic knowledge.
\newblock In {\em Proceedings of the 16th international conference on World
  Wide Web}, pages 697--706. ACM, 2007.

\bibitem[\protect\citeauthoryear{W3C}{2016}]{W3C-nary}
{Defining N-ary Relations on the Semantic Web}.
\newblock \url{http://www.w3.org/TR/swbp-n-aryRelations/}, 2016.
\newblock Accessed: 2016-01-20.

\bibitem[\protect\citeauthoryear{Wang \bgroup \em et al.\egroup
  }{2014}]{transH}
Zhen Wang, Jianwen Zhang, Jianlin Feng, and Zheng Chen.
\newblock Knowledge graph embedding by translating on hyperplanes.
\newblock In {\em Proceedings of the Twenty-Eighth AAAI Conference on
  Artificial Intelligence}, pages 1112--1119. Citeseer, 2014.

\bibitem[\protect\citeauthoryear{Xiong and Callan}{2015a}]{KG-APP2}
Chenyan Xiong and Jamie Callan.
\newblock Esdrank: Connecting query and documents through external
  semi-structured data.
\newblock In {\em Proceedings of the 24th ACM International on Conference on
  Information and Knowledge Management}, CIKM '15, pages 951--960, New York,
  NY, USA, 2015. ACM.

\bibitem[\protect\citeauthoryear{Xiong and Callan}{2015b}]{KG-APP3}
Chenyan Xiong and Jamie Callan.
\newblock Query expansion with {F}reebase.
\newblock In {\em Proceedings of the Fifth ACM International Conference on the
  Theory of Information Retrieval}. ACM, 2015.

\end{thebibliography}

\end{document}